\documentclass[lettersize,journal]{IEEEtran}
\usepackage{amsmath,amsfonts}
\usepackage{array}
\usepackage[caption=false,font=normalsize,labelfont=sf,textfont=sf]{subfig}
\usepackage{textcomp}
\usepackage{stfloats}
\usepackage{url}
\usepackage{verbatim}
\usepackage{graphicx}
\def\BibTeX{{\rm B\kern-.05em{\sc i\kern-.025em b}\kern-.08em
    T\kern-.1667em\lower.7ex\hbox{E}\kern-.125emX}}
\usepackage{balance}

\usepackage{float}
\usepackage{multirow}
\usepackage{txfonts}
\usepackage{booktabs} 
\usepackage{caption}
\usepackage{algorithm}
\usepackage{algpseudocode}
\usepackage{soul}
\usepackage{pifont}
\usepackage{color,xcolor,colortbl}
\usepackage{url}
\usepackage{CJK}
\usepackage{makecell}
 \usepackage{arydshln}
\usepackage{longtable} 
\usepackage{colortab}  
\usepackage{dcolumn}
\usepackage{setspace}
\usepackage{diagbox}
\usepackage{footnote}
\makesavenoteenv{table}
\usepackage[misc]{ifsym}

\newlength\savedwidth

\newcommand{\tabincell}[2]{\begin{tabular}{@{}#1@{}}#2\end{tabular}}

\newcommand{\xmark}{\ding{55}}
\newcommand{\rmark}{\ding{52}}

\usepackage[colorlinks,
            linkcolor=red,
            anchorcolor=blue,
            citecolor=green]{hyperref}

\begin{document}
\title{Focal Inverse Distance Transform Maps for Crowd Localization}

\author{Dingkang Liang, Wei Xu, Yingying Zhu\Letter, and Yu Zhou\Letter
\thanks{
This research was supported by the National Key Research and Development Program of China under Grant No. 2018AAA0100400, the National Natural Science Foundation of China (62176098 and 61703049) and the Natural Science Foundation of Hubei Province of China under Grant 2019CFA022. (\Letter~Corresponding authors)

Dingkang Liang, Yingying Zhu, Yu Zhou are with Huazhong University of Science and Technology; (email: dkliang@hust.edu.cn, yyzhu@hust.edu.cn, yuzhou@hust.edu.cn)

Wei Xu is with Beijing University of Posts and Telecommunications; (email: xuwei2020@bupt.edu.cn)}

}

\maketitle

\begin{abstract}

In this paper, we focus on the crowd localization task, a crucial topic of crowd analysis. Most regression-based methods utilize convolution neural networks (CNN) to regress a density map, which can not accurately locate the instance in the extremely dense scene, attributed to two crucial reasons: 1) the density map consists of a series of blurry Gaussian blobs, 2) severe overlaps exist in the dense region of the density map. To tackle this issue, we propose a novel Focal Inverse Distance Transform (FIDT) map for the crowd localization task. Compared with the density maps, the FIDT maps accurately describe the persons' locations without overlapping in dense regions. Based on the FIDT maps, a Local-Maxima-Detection-Strategy (LMDS) is derived to effectively extract the center point for each individual. Furthermore, we introduce an Independent SSIM (I-SSIM) loss to make the model tend to learn the local structural information, better recognizing local maxima. Extensive experiments demonstrate that the proposed method reports state-of-the-art localization performance on six crowd datasets and one vehicle dataset. Additionally, we find that the proposed method shows superior robustness on the negative and extremely dense scenes, which further verifies the effectiveness of the FIDT maps. The code and model will be available at \url{https://github.com/dk-liang/FIDTM}.
\end{abstract}


\section{Introduction}
\label{Introduction}

\begin{figure}[t]
	\begin{center}
		\includegraphics[width=0.96\linewidth]{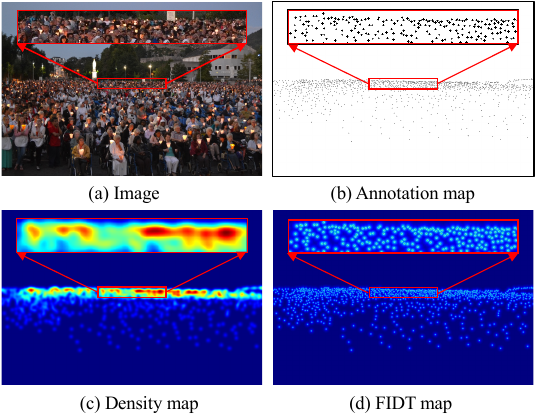}
	\end{center}
	\caption{The localization advantages of the FIDT map. (a) Input image, which has heavy occlusions and cluttered backgrounds. (b) The image only provides point-level annotations. (c) A series of Gaussian blobs represent the density map and usually accumulate density values in the dense region, making the person's location indistinguishable. (d) FIDT map uses the nearest neighbor distance information to represent each person's location, and nearby heads remain distinguishable even in dense regions.}
	\label{fig:intro1}
\end{figure}

Crowd analysis contains many sub-tasks, such as crowd detection~\cite{zhang2020attribute}, crowd counting~\cite{li2018csrnet, jiang2020density}, and crowd localization~\cite{han2021ldc, sam2020locate}. Specifically, the crowd detection task is to detect persons based on bounding boxes, an expensive way of labeling. The crowd counting aims to estimate a density map and give the total count of a crowd scene based on point-level annotations. In this paper, we focus on crowd localization, predicting a point for each person's head only based on point-level annotations, which is a more complex task compared with crowd detection and crowd counting.

The deep-learning-based detectors~\cite{ren2015faster, hu2017finding} predict the bounding box for each instance, encountering difficulties under highly congested scenes~\cite{gao2020nwpu}. 
In general, annotating the bounding box for each person in the dense crowd is expensive and laborious, so most current crowd datasets~\cite{zhang2016single,idrees2018composition} only provide point-level annotations (Fig.~\ref{fig:intro1}(b)), making the detectors~\cite{ren2015faster, hu2017finding} untrainable. Current regression-based methods~\cite{zhang2016single,sam2017switching,li2018csrnet,jiang2020density} regress a density map and output the count by integrating over the density map. However, these regression-based methods can not provide individual location and size, mainly because the Gaussian blobs of the widely used density map overlap in dense regions, making the local maxima unequal to the head locations.
However, the location and size information also play an essential role in many high-level applications, such as pedestrian tracking~\cite{li2019joint} and crowd analysis~\cite{li2018deep,zhang2016data}. To tackle the problem, PSDDN~\cite{liu2019point} and LSC-CNN~\cite{sam2020locate} utilize similar nearest-neighbor head distances to initialize the pseudo ground truth (GT) bounding boxes in a detection-like model. Essentially, both of them use bounding boxes for the training phase, still applying a complex detection framework (\textit{e.g.,} Faster R-CNN).~Actually, the pseudo GT boxes do not reflect the real head size well, leading to poor performance.

Alternatively, some methods focus on designing an appropriate map to cope with crowd localization, such as binary-like maps~\cite{abousamra2020localization,liu2019recurrent,gao2020learning} and segmentation-like maps~\cite{arteta2016counting, xu2022autoscale}. Among them, both trimap~\cite{arteta2016counting} and distance label map~\cite{xu2022autoscale} need to set the threshold of distance and number of the label in handcraft, which is empirical. Topological maps~\cite{abousamra2020localization} and IIM~\cite{gao2020learning} still apply box-level annotations in some challenge datasets (\textit{e.g.,} JHU-Crowd++ and NWPU-Crowd), which limits its application to the real world. During the testing phase, these methods regard the connected components of predicted maps as the head location. However, they easily fail in the congested scenes because the adjacent connected components may be linked together in dense regions. In other words, it is possible to incorrectly predict many heads as one.

Different from the above methods, this paper proposes a novel label named Focal Inverse Distance Transform (FIDT) map for the localization task, which provides precise location information for each person. 
It is well known that the widely used density map is blurry and indistinguishable due to each head annotation filtered with a Gaussian kernel, as shown in Fig.~\ref{fig:intro1}(c). In contrast, the proposed FIDT map is discriminative without any overlaps between nearby heads, even in extremely dense crowds, as shown in Fig.~\ref{fig:intro1}(d). In the proposed FIDT map, the closer pixels are to the head center, the higher responses they will have, which means the local maxima are equal to the head centers. Accordingly, the counting result is equal to the number of local maxima.

In FIDT maps, a local maximum represents an individual instance, so detailed local structural information can help locate the FIDT maps' local maxima. A straightforward way is to utilize the SSIM loss to improve the similarity between the predicted FIDT map and the ground truth map. However, in the FIDT map, the background's pixel value is close to 0, without structure information. The traditional SSIM loss may cause high responses for the background, which may produce false local maxima. Thus, we introduce the Independent SSIM (I-SSIM) loss to further improve the model's ability to enhance the structure information of local maxima and reduce the false local maxima in background regions.

As we mentioned above, for a given FIDT map, we can obtain the heads' position by localizing the local maxima, so the final key step is how to extract the local maxima of FIDT maps. In this paper, we propose simple yet effective post-processing named Local-Maxima-Detection-Strategy (LMDS), implemented by a simple max-pooling layer with an adaptive threshold. Furthermore, the proposed LMDS can help to classify the negative samples (\textit{e.g.,} Terra-Cotta Warriors images).

In summary, this work contributes to the following:
\begin{enumerate}
\item In order to effectively cope with the crowd localization task in dense scenes, we propose the FIDT maps. The local maxima of FIDT maps represent exact persons' locations.
\item We introduce the I-SSIM loss to make the model focus on the independent regions, enhancing the model's ability to handle the local maxima and background regions.
\item Based on the FIDT map, we design a Local-Maxima-Detection-Strategy, LMDS, which can effectively locate the predicted local maxima (head centers).
\item Extensive experiments demonstrate that the proposed method achieves state-of-the-art localization performance. Additionally, our method is robust for the negative and extremely dense scenes.
\end{enumerate}

\section{Related Works}
\subsection{Crowd analysis}
Current crowd analysis methods mainly focus on the counting task, which usually adopts CNN to regress the density maps. And the total count is obtained by integrating the density maps. Some methods work on multi-layer or multi-scale feature fusion~\cite{sindagi2019multi,jiang2019crowd,jiang2020attention} to improve the quality of predicted density maps. Some methods~\cite{zhang2019relational,zhang2019attentional,liu2019adcrowdnet} incorporate the attention mechanism into the framework, which effectively attends to the foreground regions. Multi-head layers~\cite{mazzeo2020mh} are useful that can effectively aggregate features from the conv-backbone. Using different density map representations~\cite{shi2019revisiting,wan2020kernel} is also an essential procedure in the training phase, which can promote the model's ability. Some methods make efforts to minimize the expensive labeling work in a semi-supervised~\cite{meng2021spatial,liu2020semi,xu2021crowd} or weakly-supervised~\cite{liang2022transcrowd,yang2020weakly} paradigm. Unfortunately, these counting methods only give the total count or coarse density map, which can not provide the precise position of each head, limiting the application in the real-world.

Recently, crowd localization, aiming to predict the precise position of each person's head, has been a hot topic in crowd analysis.
The deep-learning-based detectors~\cite{ren2015faster,redmon2016you,liu2016ssd} rely on bounding box annotations, which is impractical in the dense crowd due to expensive labeling costs. To address this problem, some approaches~\cite{liu2019point,sam2020locate,wang2021self} attempt to initialize the pseudo GT bounding boxes from the point-level annotations, applying the two-stage detection framework. However, the generated pseudo GT bounding boxes do not reflect the actual head sizes well, leading to unsatisfactory performance.
CL~\cite{idrees2018composition} finds the local maxima of the predicted density map with a small Gaussian kernel. Several methods attempt to predict a location map as a binary-like~\cite{abousamra2020localization,liu2019recurrent,gao2020learning} or segmentation-like map~\cite{xu2022autoscale,laradji2018blobs}. Specifically, Xu~\textit{et al.}~\cite{xu2022autoscale} propose the distance label map, which formulates the problem as a segmentation task. Shahira \textit{et al.}~\cite{abousamra2020localization} generate the binary mask by thresholding the topological map.
A recent work~\cite{gao2020learning} proposes Independent Instance Maps (IIM), and a differentiable Binarization Module is used to learn adaptive thresholds for different heads. Both topological map~\cite{abousamra2020localization} and IIM~\cite{gao2020learning} still need box-level annotations when facing challenging datasets (\textit{e.g.,} NWPU-Crowd~\cite{gao2020nwpu}).
In general, these methods usually obtain the head position by detecting the connected components of predicted maps. However, in dense regions, the connected component may be linked together, which is possible to incorrectly predict many heads as one. 

Unlike the above localization methods, we propose a new label named FIDT map. This non-overlap map utilizes the local maxima to represent persons' heads, \textit{i.e.,} the closer pixels are to the head center, the higher responses they will have. 

\subsection{Loss function}
Most crowd counting methods apply MSE as the loss function. However, only using MSE loss will cause blur, and lose the local structure information~\cite{cao2018scale}. To this end, some methods focus on designing an appropriate loss function to promote the model's ability. Specifically, BL~\cite{ma2019bayesian} regards the density map as a probability map, calculating the expected count of pixels. SPANet~\cite{cheng2019learning} proposes Maximum Excess over Pixels (MEP) loss by finding the pixel-level subregion with the highest discrepancy with ground truth. DM-Count~\cite{wang2020distribution} uses Optimal Transport (OT) to measure the similarity between the normalized predicted density map and the normalized ground truth density map. DSSINet~\cite{liu2019crowd} propose the DMS-SSIM loss to measure the structural similarity between the multiscale regions centered at the given pixel on an estimated density map and the corresponding regions on ground-truth. 

The above loss functions usually calculate the loss on the global level. In contrast, the proposed loss focuses on the structure information of local regions (instance level), which benefits the model better detecting the local maxima (\textit{i.e.,} head centers).

\subsection{Distance transform} 
Distance transform~\cite{rosenfeld1968distance} is a classical image processing operator applied in many deep-learning-based algorithms recently~\cite{hayder2017boundary, wang2020deep, wang2019deepflux,arteta2016counting}. Specifically, Hayder \textit{et al}.~\cite{hayder2017boundary} introduce a novel segment representation based on the distance transform. Wang \textit{et al}.~\cite{wang2020deep} present the Deep Distance Transform (DDT) for accurate tubular structure segmentation. 
In the crowd analysis, Arteta \textit{et al}.~\cite{arteta2016counting} and Xu~\cite{xu2022autoscale} \textit{et al}. propose semantic-like maps, discretizing a distance transform map by setting distance thresholds and transforming the localization task into a semantic task. To the best of our knowledge, we are the first to leverage the local maxima of such distance transform maps for regression-based crowd localization.

\begin{figure*}[t]
\centering
\includegraphics[width=0.99\linewidth]{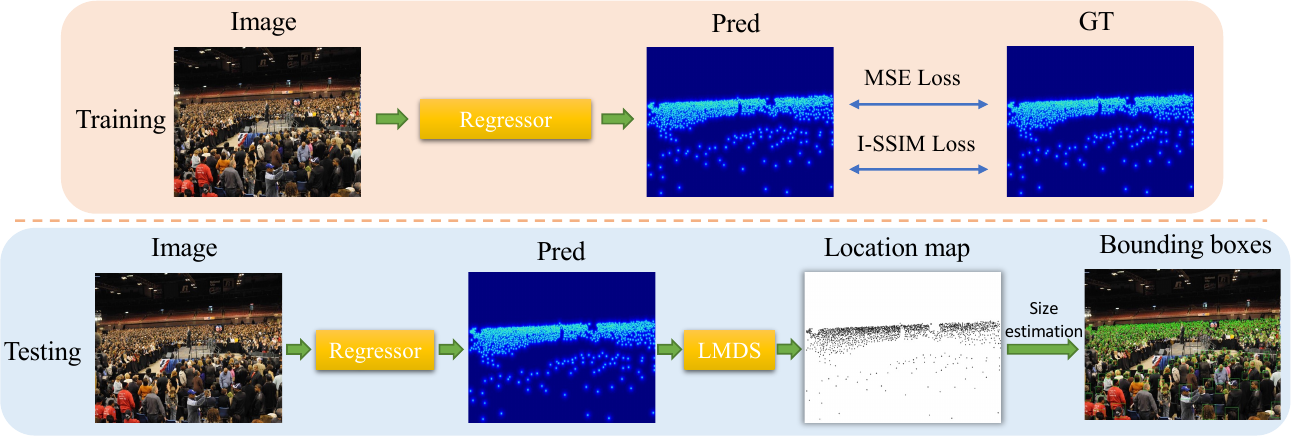}
\caption{The pipeline of our method. During the training phase, MSE loss and the proposed I-SSIM loss are adopted. During the testing phase, each person’s location can be obtained by the LMDS, and the final count is equal to the number of local maxima. Additionally, the bounding boxes can be obtained through the size estimation step.}
\centering
\label{fig:pipeline}
\end{figure*}

\begin{figure}[t]
	\begin{center}
		\includegraphics[width=0.96\linewidth]{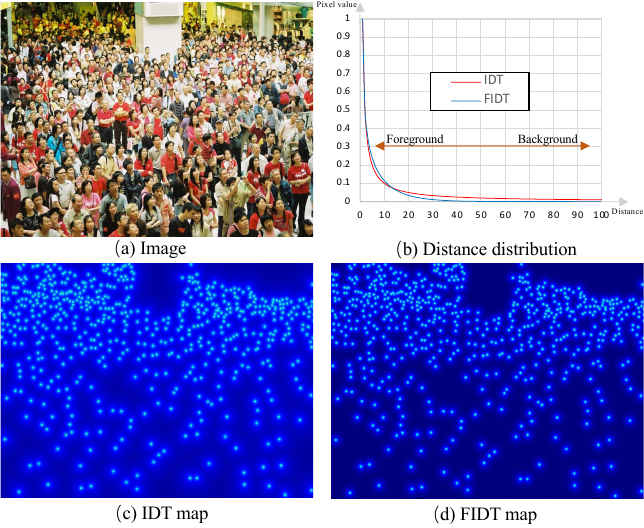}
	\end{center}

	\caption{(a) Original image. (b) The distance distribution of IDT map and FIDT map. (c) IDT map shows faster response decay from the head center and keeps high response in the background. (d) FIDT map shows slower response decay from the head center and keeps low response (close to 0) in the background.}
	\label{fig:fidt_idt}
\end{figure}

\section{Our Method}

The overview of our method is shown in Fig.~\ref{fig:pipeline}. At the training stage, a regressor is used to generate the predicted FIDT map. The MSE loss and the proposed I-SSIM loss are used to measure the difference between prediction results and ground truth. At the testing stage, a predicted FIDT map is generated, and the location map is obtained by the proposed Local-Maxima-Detection-Strategy (LMDS). Furthermore, we can obtain the bounding boxes for better visualization by a simple KNN strategy.

\subsection{Focal Inverse Distance Transform Map}  
\label{sec:fidt}

Here, we illustrate the formulation of the Euclidean distance transform map first, which is defined as:
\begin{equation}
P\left (x,y \right)=\underset{\left ( x^{'},y^{'} \right )\in B}{\min}\sqrt{\left ( x-x^{'} \right )^2+\left ( y-y^{'} \right )^2},
\label{eq:distance_transform}
\end{equation}
where $B$ represents the set of all annotations.
For an arbitrary pixel $(x,y)$, Eq.~\ref{eq:distance_transform} means that the pixel value $P(x,y)$ represents the distance between the pixel and its nearest head position (annotation). It is difficult to directly regress the distance transform map, mainly due to the large distance variations (range from 0 to the length of the image). A way is to use the inverse function to refrain from the distance variations. Specifically, the Inverse Distance Transform (IDT) map is generated, defined as:
\begin{equation}
I^{'}=\frac{1}{P\left (x, y \right ) + C},
\label{eq:idt_map}
\end{equation}
where $I'$ is the IDT map, and $ C $ is an additional constant (set $ C $  = 1) to avoid being divided by zero, as the range of distance transform map is [0, $+\infty$). The IDT map is a special form of the iKNN map~\cite{olmschenk2019improving} (when the K is set as 1). It is noteworthy that \cite{olmschenk2019improving} only uses the iKNN map for the counting task instead of the localization task. Compared to the widely used density map, the IDT (i1NN~\cite{olmschenk2019improving}) map can accurately represent the individual locations, which correspond to the local maxima. However, the IDT map presents a faster response decay away from the head center and slower response decay in the background, as shown in the distance distribution curve in Fig.~\ref{fig:fidt_idt}(b). Ideally, the decay should be slower away from the head, and the response of the background should be quickly close to 0, which means the model should focus on the foregrounds (head regions).
Thus, we propose the Focal Inverse Distance Transform (FIDT) map, defined as:
\begin{equation}
I= \frac{1}{P\left ( x,y \right )^{\left ( \alpha \times P{\left ( x,y \right )}+\beta  \right )}+C},
\label{eq:fidt_map}
\end{equation}
where $I$ is the FIDT map we proposed, $\alpha$ and $\beta$ set as 0.02 and 0.75, respectively. As shown in Fig.~\ref{fig:fidt_idt}(b), the curve examples of the IDT and FIDT map are illustrated. Compared with the IDT map, the FIDT map shows slower response decay away from the head center, and the response of background is close to 0, as shown in Fig.~\ref{fig:fidt_idt}(c) and Fig.~\ref{fig:fidt_idt}(d). It is noteworthy that the proposed FIDT map is totally different from the density map that uses small Gaussian kernels. The latter still presents overlap in extremely dense scenes, and a recent method~\cite{xu2022autoscale} has demonstrated that small Gaussian kernels can not report satisfying localization and counting performance.

\begin{figure}[t]
\centering
\begin{minipage}{1\linewidth}
\begin{algorithm}[H]
\caption{Local Maxima Detection Strategy (LMDS)}
\label{alg:extract_position}
\begin{algorithmic}[1]
    \State \textbf{Input:} Predicted FIDT map
    \State \textbf{Output:} The coordinates of the persons and the total count
    \Function {Extract$\_$Position}{$input$} 
        \State {$pos\_ind = maxpooling(input, size = (3,3))$}
        \State {$pos\_ind = (pos\_ind == input)$}
        \State $matrix = pos\_ind \times input$
        \If{$max(matrix)<T_f$}
        \State $count=0$
        \State $coordinates = None$
        \Else 
        \State $T_a = 100/ 255.0  \times max(matrix)$
        \State $matrix[matrix \textless T_a] = 0$
        \State $matrix[matrix \textgreater 0] = 1$
        \State $count = sum(matrix)$
        \State $coordinates = nonzeros(matrix)$
        \EndIf
        \State \Return {$count, coordinates$} 
    \EndFunction  
\end{algorithmic}
\end{algorithm}
\end{minipage}
\end{figure}

\subsection{Localization framework}  
\subsubsection{Regressor}

To verify the effectiveness of the proposed FIDT maps, we use a straightforward base network to regress the FIDT maps. The corresponding individual center of the FIDT map is equal to the local maximum, so high-resolution representations are essential. Here, following IIM~\cite{gao2020learning}, we use HRNET~\cite{wang2020hrnet} as the base network, and we add one convolution and two transposed convolution layers as the representation head based on the HRNET~\cite{wang2020hrnet}. Note that the regressor can be replaced by any crowd regressor, such as CSRNET~\cite{li2018csrnet}, BL~\cite{ma2019bayesian}.

\subsubsection{Local Maxima Detection Strategy}

Given a predicted FIDT map, we can obtain the persons' positions by localizing the local maxima. We call this process as Local-Maxima-Detection-Strategy (LMDS), as illustrated in Algorithm~\ref{alg:extract_position}. Specifically, we first utilize a $3\times3$ max-pooling to obtain all local maxima (candidate points). However, these candidate points may contain some false positives from the background. We observe that the pixel values of true positives are much larger than the pixel values of false positives, which means a local maximum is likely to be a person if its pixel value is large enough. This inspires us to utilize an adaptive threshold $T_a$ to filter the false positives. Thus, given a series of candidate points $M$, the final selected points are those whose values are no less than $T_a$, which is equal to $100/255.0$ times the maximum of $M$. Recent dataset~\cite{ gao2020nwpu} provides some negative samples, which consists of some scenes without persons and is similar to crowd scenes. We can not judge whether the original images contain persons based on the predicted density maps. However, given a predicted FIDT map, if the maximum of $M$ is smaller than a tiny fixed threshold $T_f$ (set as 0.10), this means the input image is a negative sample, and the LMDS will set the counting result as 0. An example of obtained location map is shown in Fig.~\ref{fig:pipeline}.

Although the real individual sizes are not provided, we can generate pseudo individual bounding boxes from the predicted FIDT map. Here, the bounding box is a square surrounding a head.
Once we get the predicted FIDT map, we first extract the coordinates of the head centers, which can be implemented efficiently using the proposed LMDS. Then, we estimate the instance size, using the K-nearest neighbours distance, which is defined as:
\begin{equation}
s_{\left ( x,y \right )\in P}=\min\left\{
\begin{matrix}
 \bar{d}= f \times \frac{1}{k}\sum_{1}^{k} d_{\left ( x,y \right )}^{k}
\\
\\
\min\left (img\_w,img\_h  \right )\times 0.05
\end{matrix}\right.
\label{eq:boxgenarate}
\end{equation}
where $s_{(x,y)}$ means the size of the instance, which locate in $(x,y)$, and $P$ is the set of predicted head positions. $\bar{d}$ is the average distance, which is calculated between point $P_{(x,y)}$ and its K-nearest neighbours, and a scalar factor $f$ is used to restrain the size. In very sparse regions, the $\bar{d}$ may be bigger than the real size of persons, so we choose a threshold related to image size to restrain the object size, as described in Eq.~\ref{eq:boxgenarate}. Note that Eq.~\ref{eq:boxgenarate} is only used in the testing phase for visualizations, and the size of bounding boxes does not influence the localization performance.

\begin{table*}[t]
\centering
\small
\setlength{\tabcolsep}{1.5mm}
\caption{Quantitative comparison of the localization performance on the NWPU-Crowd dataset. The results of other methods are from the online benchmark website~\cite{gao2020nwpu}. F, P, and R refer to the F-measure, precision and recall, respectively.}
\label{tab:nwpu_loc}

	\begin{tabular}{ccccccccccccccc}
		\toprule
		\multirow{4}{*}{Method}	 &\multirow{4}{*}{\tabincell{c}{Training\\Labels}} &\multicolumn{6}{c}{Validation set}  &\multicolumn{6}{c}{Test set}  \\ %
		\cmidrule(r){3-8} \cmidrule{9-14}
		  &&\multicolumn{3}{c}{$\sigma_l$} &\multicolumn{3}{c}{$\sigma_s$}  &\multicolumn{3}{c}{$\sigma_l$}  &\multicolumn{3}{c}{$\sigma_s$}   \\
		\cmidrule(r){3-8} \cmidrule{9-14}
		  &&F (\%)&P (\%)&R (\%)  & F (\%) &P (\%)&R (\%) & F (\%)&P (\%)&R (\%)& F (\%)&P (\%)&R (\%)\\
		\cmidrule(r){1-2} \cmidrule(r){3-8} \cmidrule{9-14}
		Faster RCNN~\cite{ren2015faster}   &Box  &7.3&\textbf{96.4}&3.8 &6.8&\textbf{90.0}&3.5&6.7&\textbf{95.8}&3.5&6.3&\textbf{89.4}&3.3\\
		TinyFaces~\cite{hu2017finding}    &Box  &59.8&54.3&66.6 &55.3&50.2&61.7 &56.7&52.9&61.1 &52.6&49.1&56.6    \\
		TopoCount~\cite{abousamra2020localization}  &Box    &-&-&-  &-&-&- &{69.1}&69.5&68.7 &60.1&60.5&59.8 \\
		\cmidrule(r){1-2} \cmidrule(r){3-8} \cmidrule{9-14}
		VGG+GPR~\cite{gao2019domain}   &Point   &56.3&61.0&52.2  &46.0&49.9&42.7 &52.5&55.8&49.6 &42.6&45.3&40.2 \\
		RAZ\_Loc~\cite{liu2019recurrent}   &Point  &62.5&69.2&56.9 &54.5&60.5&49.6  &59.8&66.6&54.3  &51.7&57.6&47.0  \\
		Crowd-SDNet~\cite{wang2021self} &Point &-&-&-&-&-&-&63.7&65.1&62.4&-&-&-\\
	    AutoScale$^{*}$~\cite{xu2022autoscale} &Point &66.8&70.1&63.8 &60.0&62.9&57.3 &62.0&67.3&57.4 &54.4&59.1&50.4 \\
		GL~\cite{wan2021generalized} &Point  &-&-&-&-&-&-&66.0&80.0&56.2&-&-&- \\
		SCALNet~\cite{wang2021dense} &Point    &72.4&73.5&71.4  &66.9&67.9&65.9 &{69.1}&69.2&63.6 &63.6&63.7&63.6 \\
		\textbf{Ours}    &Point &\textbf{78.9}&82.2&\textbf{75.9} &\textbf{73.7}&76.7&\textbf{70.9} &\textbf{75.5}&79.7&\textbf{71.7} &\textbf{70.5}&74.4&\textbf{66.9}\\
		\bottomrule
	\end{tabular}
\end{table*}

\subsection{Independent SSIM Loss}

Just using MSE loss to supervise the training phase will cause some negative impacts, such as blur effect and losing local structure information~\cite{cao2018scale}. Some methods~\cite{cao2018scale,liu2019crowd} have proved that SSIM loss can improve the quality of the predicted map. SSIM is defined as:
\begin{equation}
SSIM(E,G)=\frac{\left ( 2\mu_{E}\mu_{G}+\lambda _{1} \right )\left ( 2\sigma_{EG}+\lambda_{2}  \right )}{\left ( \mu _{E}^{2}+\mu_{G}^{2}+\lambda_{1} \right )\left ( \sigma_{E}^2+\sigma_{G}^2 +\lambda_{2}\right )},
\label{eq:ssim}
\end{equation}
where $E$ and $G$ represent the estimated map and ground-truth map, respectively. The $\mu$ and $\sigma$ are the mean and variance. 
$\lambda_{1}$ and $\lambda_{2}$ are set to 0.0001 and 0.0009 to avoid being divided by zero. The value range of $SSIM$ is [-1,1], and $SSIM$ = 1 means the estimated map is the same as the ground truth, so the SSIM loss is defined as:
\begin{equation}
L_{S}(E,G)=1-SSIM(E,G).
\label{eq:L_ssim}
\end{equation}
In general, the SSIM loss utilizes a sliding window to scan the whole predicted map without distinguishing the foreground (head region) and background. However, for the localization task, relying on detecting the local maxima, the model should focus on local maxima. The global SSIM loss may generate high responses, causing some false local maxima in the background. Thus, we propose the Independent SSIM (I-SSIM) loss, defined as:

\begin{table}[t]
\centering
\small
\caption{Quantitative evaluation of localization-based methods on the UCF-QNRF dataset. We report the average precision, average Recall, and average F-measure at different distance thresholds (1, 2, 3, \dots, 100).
}
\label{tab:qnrf_loc}
\begin{tabular}{cccc}
\toprule
Method &Av.Precision&Av.Recall&F-measure\\
\cmidrule(r){1-1} \cmidrule(r){2-4}
MCNN~\cite{zhang2016single}&59.93\%&63.50\%&61.66\%\\
CL~\cite{idrees2018composition}&75.80\%&59.75\%&66.82\%\\
LCFCN~\cite{laradji2018blobs}&\textcolor{black}{77.89\%}&\textcolor{black}{52.40\%}&\textcolor{black}{62.65\%}\\
Method in~\cite{ribera2019}&\textcolor{black}{75.46\%}&\textcolor{black}{49.87\%} & \textcolor{black}{60.05\%}\\
LSC-CNN\cite{sam2020locate}&\textcolor{black}{74.62\%} &\textcolor{black}{73.50\%} & \textcolor{black}{74.06\%}\\
	GL~\cite{wan2021generalized} &78.20\% & 74.80\% & 76.30\% \\
TopoCount\cite{abousamra2020localization} &\textcolor{black}{81.77\%} &\textcolor{black}{78.96\%} & \textcolor{black}{80.34\%}\\
\textbf{Ours}&\textbf{84.49}\%&\textbf{80.10}\%&\textbf{82.23\%}\\
 \bottomrule
\end{tabular}
\end{table}

\begin{equation}
L_{I-S}=\frac{1}{N}\sum_{n=1}^{N}L_{S}\left ( E_{n},G_{n} \right ),
\label{eq:I_S}
\end{equation}
where $N$ means the total number of persons, $E_{n}$ and $G_{n}$ mean the estimated and ground truth for the $n$-th independent instance region, and the region size of each instance is set as $30\times30$ for all datasets, mainly because we observe that this size can contain the entire head region without redundant background for most independent instance. The final training objective $L$ is defined as below:
\begin{equation}
L =  L_{MSE} + L_{I-S},
\label{eq:L}
\end{equation}
where $L_{MSE}$ and $L_{I-S}$ refer to the MSE loss and the proposed I-SSIM loss, respectively. 

\begin{table}[t]
\small
\setlength{\tabcolsep}{1.0mm}
\centering
\caption{Quantitative evaluation of localization-based methods on the JHU-Crowd++ dataset using Precision (P), Recall (R), and F-measure (F).}
\label{tab:jhu_loc}
\begin{tabular}{ ccccccc }
 \toprule
 {\multirow{2}{*}{Method}} & \multicolumn{3}{c}{$\sigma$ = 4}&\multicolumn{3}{c}{$\sigma$ = 8} \\
\cmidrule(r){2-4} \cmidrule{5-7}
& P (\%)& R (\%) &F (\%) &P (\%) &R (\%)&F (\%) \\
\cmidrule(r){1-1} \cmidrule(r){2-4} \cmidrule{5-7}
TopoCount~\cite{abousamra2020localization} &31.5\%&28.8\%&30.1\%&\textbf{63.6\%}&58.3\%&60.8\%\\
\textbf{Ours}&\textbf{38.9\%}&\textbf{38.7\%}&\textbf{38.8\%}&62.5\%&\textbf{62.4\%}&\textbf{62.4\%}\\
\bottomrule
\end{tabular}
\end{table}

\begin{figure}[t]
\centering
\includegraphics[width=1\linewidth]{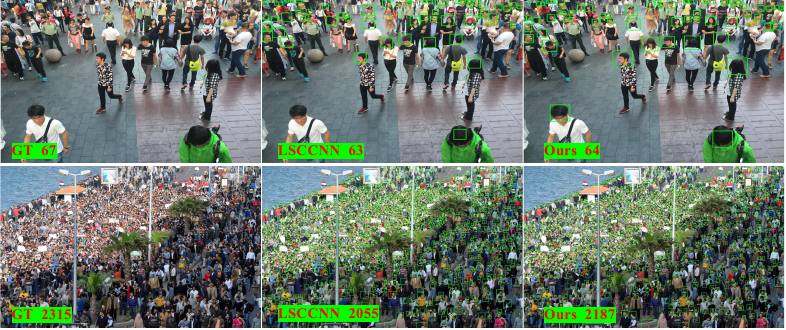}
\caption{Qualitative visualization of detected persons' locations by the proposed method. We use the proposed KNN strategy to generate bounding boxes (green boxes), compared with LSC-CNN~\cite{sam2020locate}.}
\centering
\label{fig:boxes}
\end{figure}

\section{Implement details}
We augment the training data using random cropping and horizontal ﬂipping. The crop size is 256 × 256 for Part A and Part B and 512 × 512 for other datasets. We set $k$ as 4 and $f$ as 0.1 to generate the bounding boxes. The $\alpha$ and $\beta$ set as 0.02 and 0.75 respectively. We use Adam to optimize the model with the learning rate of 1e-4, and the weight decay is set as 5e-4. We set the size of the training batch to 16. We resize the images to make sure that the longer side is smaller than 2048 for NWPU-Crowd~\cite{gao2020nwpu}, JHU-Crowd++~\cite{sindagi2020jhu-crowd++}, and UCF-QNRF~\cite{idrees2018composition} datasets. 

\subsection{Evaluation metrics}
\textbf{Localization Metrics.}
Precision, Recall, and F-measure are adopted to evaluate the performance on the NWPU-Crowd dataset, defined by~\cite{gao2020nwpu}. When the distance between the given predicted point $P_p$ and ground truth point $P_g$ is less than a distance threshold $\sigma$, it means the $P_p$ and $P_g$ are successfully matched. The $\sigma$ is related to the real head size (this dataset provides box-level annotation). Specifically, Wang \textit{et al}.~\cite{gao2020nwpu} give two thresholds:
\begin{equation}
\sigma_s = \min(w,h)/2,
\label{eq:sigma_s}
\end{equation}
\begin{equation}
\sigma_l={\sqrt {{w^2} + {h^2}}}/2,
\label{eq:sigma_l}
\end{equation}
and the former is a stricter criterion than the latter. For the UCF-QNRF dataset, similar to CL~\cite{idrees2018composition}, we calculate the Precision, Recall, and F-measure at various thresholds (1, 2, 3, . . . , 100 pixels). For JHU-Crowd++, ShanghaiTech Part A, Part B, and UCF\_CC\_50 datasets, we choose two fixed thresholds ($\sigma$ = 4, 8) for evaluation. 

\textbf{Counting Metrics.}
We use the Mean Absolute Error (MAE) and Mean Square Error (MSE) as the counting metrics, defined as:
\begin{equation}
MAE=\frac{1}{M}\sum_{i=1}^{M}\left |P_{i}-G_{i} \right|,
\label{eq:MAE}
\end{equation}
\begin{equation}
MSE=\sqrt{\frac{1}{M}\sum_{i=1}^{M}\left |P_{i}-G_{i} \right|^{2}}, 
\label{eq:MSE}
\end{equation}
where $M$ is the number of testing images, $P_{i}$ and $G_{i}$ are the predicted and ground truth count of the $i$-th image, respectively.

\subsection{Dataset} 
We evaluate our method on six challenging public datasets, each being elaborated below. 

\textbf{NWPU-Crowd}~\cite{gao2020nwpu}, a large-scale and challenging dataset, consists of 5,109 images, elaborately annotating 2,133,375 instances. The dataset provides 351 negative samples, testing the robustness of the model. 
The results are from an online evaluation benchmark website.

\textbf{JHU-CROWD++}~\cite{sindagi2020jhu-crowd++} contains 2,722 training images, 500 validation images, and 1,600 test images, collected from diverse scenarios. The total number of persons in each image ranges from 0 to 25,791.

\textbf{UCF-QNRF}~\cite{idrees2018composition} contains 1,535 images and about one million annotations. It has a count range of 49 to 12,865, with an average count of 815.4. 

\textbf{ShanghaiTech}~\cite{zhang2016single} consists of Part A and Part B with a total count of 1,198 images. In particular, Part A contains 300 training images and 182 testing images, and Part B consists of 400 training images and 316 testing images. 

\textbf{UCF\_CC\_50}~\cite{idrees2013multi} contains 50 gray images captured in extremely congested scenes. The number of crowd counts varies from 96 to 4,633. It is a challenging dataset due to the heavy background noise and the limited number of images.

\textbf{TRANCOS}~\cite{guerrero2015extremely} contains 1,244 images captured in traffic congestion situations with 46,796 annotations, providing a region of interest (ROI) for each image.

\section{Results and Analysis}

\begin{figure*}[t]
	\begin{center}
		\includegraphics[width=0.96\linewidth]{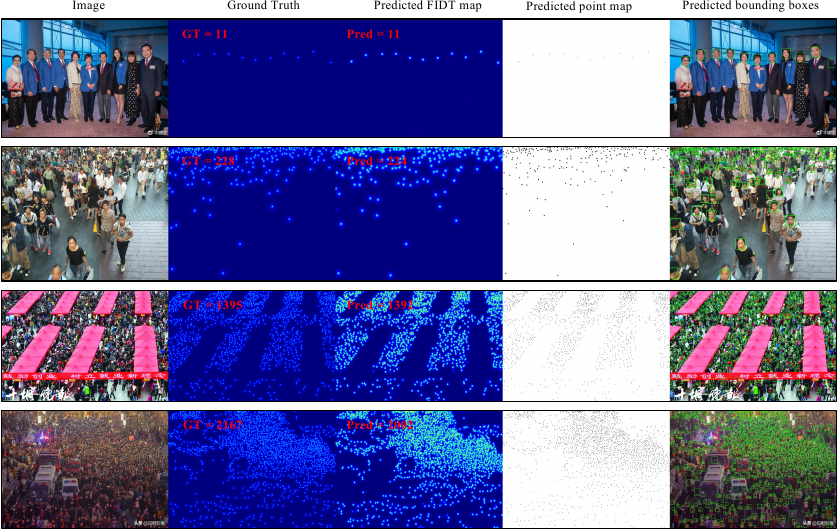}
	\end{center}
	\caption{Qualitative visualizations of our method. From left to right, there are testing images, ground truth maps, predicted FIDT maps, predicted point maps, and predicted bounding boxes.}
	\label{fig:box_visual}
\end{figure*}

\begin{table*}[t]

\small
\centering
\caption{Comparison of the localization performance on the ShanghaiTech Part A~\cite{zhang2016single} and ShanghaiTech Part B~\cite{zhang2016single} datasets using Precision (P), Recall (R), and F-measure (F).}
\label{tab:A_B_loc}
\begin{tabular}{ ccccccccccccc }
 \toprule
 {\multirow{3}{*}{Method}} &\multicolumn{6}{c}{Part A} &\multicolumn{6}{c}{Part B} \\
 \cmidrule(r){2-7} \cmidrule{8-13}
 &\multicolumn{3}{c}{$\sigma$ = 4}&\multicolumn{3}{c}{$\sigma$ = 8} &\multicolumn{3}{c}{$\sigma$ = 4}&\multicolumn{3}{c}{$\sigma$ = 8} \\
\cmidrule(r){2-7} \cmidrule{8-13}
& P (\%)& R (\%) &F (\%) &P (\%) &R (\%)&F (\%) & P (\%)& R (\%) &F (\%) &P (\%) &R (\%)&F (\%) \\
\cmidrule(r){1-1} \cmidrule(r){2-7} \cmidrule{8-13}
LCFCN\cite{laradji2018blobs}  &43.3\% &26.0\% &32.5\%& 75.1\% &45.1\% &56.3\%  &- &- &- & - &- &-\\
Method in~\cite{ribera2019}  &34.9\% &20.7\% &25.9\%& 67.7\% &44.8\%& 53.9\%  &- &- &- & - &- &-\\
LSC-CNN~\cite{sam2020locate}  &33.4\% &31.9\% & 32.6\%&63.9\% &61.0\% &62.4\%
&29.7\%&29.2\%&29.5\%&57.5\%&56.7\%&57.0\%  \\
TopoCount~\cite{abousamra2020localization} &41.7\% &40.6\% &41.1\% &74.6\% &72.7\% &73.6\%
&63.4\%&63.1\%&63.2\%&82.3\%&81.8\%&82.0\%  \\
\textbf{Ours} &\textbf{59.1\%}&\textbf{58.2\%}&\textbf{58.6\%}&\textbf{78.2}\%&\textbf{77.0\%} &\textbf{77.6\%}
&\textbf{64.9\%}&\textbf{64.5\%}&\textbf{64.7\%}&\textbf{83.9\%}&\textbf{83.2\%}&\textbf{83.5\%}
\\
\bottomrule
\end{tabular}
\end{table*}

\begin{table}[t]
\small
\setlength{\tabcolsep}{1mm}
\centering
\caption{Quantitative evaluation of localization-based methods on the UCF\_CC\_50 dataset using Precision (P), Recall (R), and F-measure (F). $\dag$ represents that the networks are trained by ourselves.}
\label{tab:ucf_50_loc}
\begin{tabular}{ ccccccc }
 \toprule
 {\multirow{2}{*}{Method}} & \multicolumn{3}{c}{$\sigma$ = 4}&\multicolumn{3}{c}{$\sigma$ = 8} \\
\cmidrule(r){2-4} \cmidrule{5-7}
& P (\%)& R (\%) &F (\%) &P (\%) &R (\%)&F (\%) \\
\cmidrule(r){1-1} \cmidrule(r){2-4} \cmidrule{5-7}
LSC-CNN\dag~\cite{sam2020locate} & 37.7\%&39.5\%&38.6\%& 57.8\%&61.1\%&59.4\%\\
AutoScale\dag~\cite{xu2022autoscale} & 37.8\%&40.5\%&39.1\%& 59.0\%&62.3\%&60.6\%\\
TopoCount\dag~\cite{abousamra2020localization} & 39.5\% &42.0\%&40.7\%& 62.5\%&66.9\%&64.6\%\\
\textbf{Ours} & \textbf{46.5\%}&\textbf{49.0\%}& \textbf{47.7\%} & \textbf{67.0\%}&\textbf{70.6\%}&\textbf{68.7\%}\\
\bottomrule
\end{tabular}
\end{table}

\subsection{Crowd localization} 
Tab.~\ref{tab:nwpu_loc},~\ref{tab:qnrf_loc},~\ref{tab:jhu_loc},~\ref{tab:A_B_loc}, and~\ref{tab:ucf_50_loc} compare the localization performance of the proposed method against the state-of-the-art methods. The results of other methods~\cite{sam2020locate, abousamra2020localization, xu2022autoscale} are from the official code and model, and we directly utilize their predicted coordinates for evaluation. The evaluated localization code is provided by~\cite{gao2020nwpu}. 

The results of the NWPU-Crowd dataset are from an online benchmark website, making sure to evaluate the localization performance fairly. As shown in Tab.~\ref{tab:nwpu_loc}, we can observe that the proposed method outperforms the popular detectors, including Faster RCNN~\cite{ren2015faster} and TinyFaces~\cite{hu2017finding}, by a significant margin. Compared with SCALNet~\cite{wang2021dense} and TopoCount~\cite{abousamra2020localization}, our method outperforms them by at least 6.4\% for $\sigma_l$ (6.9\% for $\sigma_s$) F-measure. Note that TopoCount~\cite{abousamra2020localization} still applies the box-level annotations for training on the NWPU-Crowd dataset, while our method just utilizes the point-level.

For the dense dataset UCF-QNRF, as shown in Tab.~\ref{tab:qnrf_loc}, the proposed method reports the highest Precision and Recall. 
For the JHU-Crowd++ dataset, as depicted in Tab.~\ref{tab:jhu_loc}, the proposed method improves the state-of-the-art method TopoCount~\cite{abousamra2020localization} by 8.7\% F-measure for the very strict setting $\sigma = 4$. 

For the two sparse datasets, ShanghaiTech Part A and Part B, as depicted in Tab.~\ref{tab:A_B_loc}, the proposed method improves the TopoCount~\cite{abousamra2020localization} by 17.5\% F-measure for the stricter setting $\sigma$ = 4 on part A, and 1.5\% F-measure on part B. It indicates that the proposed method can effectively cope with dense and sparse scenes. 

For the gray images, UCF\_CC\_50 dataset (Tab.~\ref{tab:ucf_50_loc}), our method surpasses the other localization methods by a significant margin, \textit{i.e.,} more than 7\% F-measure improvement on the $\sigma = 4$. This impressive result demonstrates that our method is robust to the degraded images. 

Additionally, we qualitatively evaluate the proposed method by visualizing the bounding boxes on the various crowd scenes in Fig.~\ref{fig:boxes} and Fig.~\ref{fig:box_visual}. The proposed method gives competitive bounding boxes compared with LSC-CNN~\cite{sam2020locate} and achieves impressive localization performance under various crowd scenes. It is noteworthy that the bounding boxes are only used for visualizations during the testing phase, and the size of bounding boxes does not affect the localization performance.

\begin{table*}[t]
    \small
	\setlength{\tabcolsep}{3mm}
	\centering
	\caption{Comparison of the counting performance on the NWPU-Crowd. $S0\!\sim\!S4$ respectively indicate five categories according to the different number range: $0$, $(0, 100]$, $(100, 500]$, $(500, 5000]$, $\textgreater5000$. * means the localization-based methods, which can provide the position information.}
    \label{tab:NWPU_counting}

		\begin{tabular}{ccccccccc}
			\toprule
			\multirow{4}{*}{Method}&\multirow{4}{*}{\tabincell{c}{Output\\Position\\Information}}&\multicolumn{2}{c}{validation set}&\multicolumn{4}{c}{Test set}     \\
			\cmidrule(r){3-4} \cmidrule{5-8} 
			&&\multicolumn{2}{c}{Overall}&\multicolumn{2}{c}{Overall} &\multicolumn{3}{c}{  Scene Level (only MAE)} \\
			\cmidrule(r){3-4} \cmidrule{5-8}
			&& MAE &MSE  & MAE &MSE  &Avg. & $S0 \sim S4$   \\
			\cmidrule(r){1-2} \cmidrule(r){3-4} \cmidrule{5-8}
			C3F-VGG~\cite{gao2019c}  &\xmark&105.79 &504.39& 127.0 & 439.6  & {666.9} & 140.9/26.5/58.0/307.1/2801.8 \\
			CSRNet~\cite{li2018csrnet}  &\xmark&104.89 &433.48&121.3 & 387.8 & 522.7 & 176.0/35.8/59.8/285.8/2055.8  \\
			CAN~\cite{liu2019context}  &\xmark&93.58 &489.90& {106.3} & \textbf{386.5}  &612.2 & 82.6/14.7/46.6/269.7/2647.0  \\
			SCAR~\cite{gao2019scar}  &\xmark&81.57 &397.92&110.0 & 495.3  & 718.3 & 122.9/16.7/46.0/241.7/3164.3    \\
			BL~\cite{ma2019bayesian}  &\xmark&93.64 &470.38&105.4  &454.2   & 750.5 & 66.5/8.7/41.2/249.9/3386.4\\
			SFCN~\cite{wang2019learning}  &\xmark&95.46 &608.32& 105.7 & 424.1  & 712.7 & 54.2/14.8/44.4/249.6/3200.5\\
	        KDMG~\cite{wan2020kernel} &\xmark&-&-&100.5&415.5&632.7&77.3/10.3/38.5/259.4/2777.9\\
			NoisyCC~\cite{wan2020modeling}   &\xmark  &-&-&96.9&534.2&608.1&218.7/10.7/35.2/203.2/2572.8 \\
			DM-Count ~\cite{wang2020distribution}  &\xmark &\textbf{70.5}&\textbf{357.6}&\textbf{88.4}&388.6&\textbf{498.0}&146.6/7.6/31.2/228.7/2075.8 \\
			\cmidrule(r){1-2} \cmidrule(r){3-4} \cmidrule{5-8}
		    RAZ\_loc*~\cite{liu2019recurrent}&\rmark&128.7&665.4&151.4& 634.6& 1166.0 & 60.6/17.1/48.3/364.7/5339.0   \\
		    AutoScale*~\cite{xu2022autoscale} &\rmark&97.3&571.2&123.9& 515.5& 871.0 & 42.3/18.8/46.1/301.7/3947.0 \\
			TopoCount*~\cite{abousamra2020localization}  &\rmark&-&-&107.8&438.5&-&- \\
			SCALNet*~\cite{wang2021dense} &\rmark &64.4&251.1&86.8&339.9&429.5&92.0/11.2/41.1/227.7/1775.3\\
			\textbf{Ours*}  &\rmark& \textbf{51.4}&\textbf{107.6}&\textbf{86.0} & \textbf{312.5} & \textbf{390.6} & 21.6/13.7/55.6/217.1/1645.4 \\
			\bottomrule
		\end{tabular}

\end{table*}

\begin{table*}[!t]
\centering
\small
\caption{Comparison of the counting performance on the JHU-Crowd++, UCF-QNRF, ShanghaiTech Part A, Part B and UCF\_CC\_50 datasets. * means the localization-based methods, which can provide the position information.
}
\label{tab:qnrf_A_B_counting}
\setlength{\tabcolsep}{2.5mm}{
\begin{tabular}{lccccccccccc}
 \toprule
 {\multirow{3}{*}{Method}} &{\multirow{3}{*}{\tabincell{c}{Output\\Position\\Information}}}&\multicolumn{2}{c}{\multirow{2}*{JHU}}&\multicolumn{2}{c}{\multirow{2}*{QNRF}}   &\multicolumn{2}{c}{\multirow{2}*{Part A}} &\multicolumn{2}{c}{\multirow{2}*{Part B}} &\multicolumn{2}{c}{\multirow{2}*{UCF\_CC\_50}}\\
&&&&&&&&&\\
\cmidrule(r){3-4} \cmidrule(r){5-6} \cmidrule(r){7-8} \cmidrule(r){9-10} \cmidrule(r){11-12}   
&& MAE & MSE& MAE & MSE &MAE&MSE &MAE&MSE &MAE&MSE\\
\cmidrule(r){1-2} \cmidrule(r){3-4} \cmidrule(r){5-6} \cmidrule(r){7-8} \cmidrule(r){9-10} \cmidrule(r){11-12}   
     CSRNet~\cite{li2018csrnet}                 &\xmark & 85.9& 309.2  &-&-            &68.2&115.0   & 10.6&16.0 & 266.1 & 397.5\\
    SFCN \cite{wang2019learning} &\xmark & 77.5& 297.6&102.0&171.4&64.8&107.5&7.6&13.0 & 214.2& 318.2\\
     L2SM~\cite{xu2019learn}                     &\xmark &-&-  &104.7 &173.6  &64.2&98.4   &7.2 &11.1 & 188.4 & 315.3\\ 
     CG-DRCN \cite{sindagi2019pushing}&\xmark &82.3& 328.0&112.2&176.3&64.0&98.4&8.5&14.4 & - & -\\
     MUD-iKNN \cite{olmschenk2019improving} &\xmark & -&-&104.0&172.0&68.0&117.7&13.4&21.4 & 237.7 & 305.7\\
     DSSI-Net \cite{liu2019crowd}&\xmark & 133.5  & 416.5&99.1&159.2&60.6&96.0&6.9&10.3 & 216.9& 302.4\\
     MBTTBF \cite{sindagi2019multi} &\xmark & 81.8&299.1&97.5&165.2&60.2&94.1&8.0&15.5 & 233.1 & 300.9 \\
     BL~\cite{ma2019bayesian}  &\xmark&75.0& 299.9  &88.7&154.8                       &62.8&101.8   & 7.7 & 12.7 & 229.3 & 308.2\\
     RPNet~\cite{yang2020reverse} &\xmark &-&-  &- &-                         &61.2&96.9    &8.1 &11.6 & - & -\\
     ASNet~\cite{jiang2020attention}  &\xmark &-&-   &91.6 &159.7           &57.8&\textbf{90.1}    &- &-  & \textbf{174.8} & \textbf{251.6}\\
     AMSNet~\cite{hu2020count} &\xmark &-&-  &101.8&163.2                  &56.7&93.4    &\textbf{6.7}&\textbf{10.2} & 208.6 & 296.3\\
     LibraNet~\cite{liu2020weighing} &\xmark &-&-  &88.1 & \textbf{143.7}&\textbf{55.9}&97.1      &7.3&11.3 & 181.2 & 262.2\\
    KDMG~\cite{wan2020kernel}   &\xmark&69.7&268.3 &99.5&173.0&63.8&99.2&7.8&12.7 & - & -\\
     NoisyCC~\cite{wan2020modeling}   &\xmark&\textbf{67.7}&\textbf{258.5}  &85.8&150.6&61.9&99.6&7.4&11.3 & - & -\\
     DM-Count ~\cite{wang2020distribution}  &\xmark&-&-   &\textbf{85.6}&148.3&59.7&95.7&7.4&11.8  & 211.0 & 291.5\\
     \cmidrule(r){1-2} \cmidrule(r){3-4} \cmidrule(r){5-6} \cmidrule(r){7-8} \cmidrule(r){9-10} \cmidrule(r){11-12}   
     RAZ\_loc+*~\cite{liu2019recurrent} &\rmark  &-&-   &118.0&198.0  &71.6& 120.1 & 9.9 & 15.6   & - & -\\
     PSDDN*~\cite{liu2019point}                 &\rmark  &-&-   &-&- &65.9&112.3&9.1&14.2 & 359.4 & 514.8\\
     LSC-CNN*~\cite{sam2020locate} &\rmark &112.7&454.4   &120.5&218.2 &66.4&117.0  &8.1 &12.7 & 225.6 & 302.7\\
     Crowd-SDNet*~\cite{wang2021self}&\rmark &-&-&-&-&65.1&104.4&7.8&12.6 & - & -\\
     AutoScale*~\cite{xu2022autoscale}&\rmark &85.6&356.1&104.4&174.2&65.8&112.1&8.6&13.9 & 210.5 & 287.4\\
     TopoCount*~\cite{abousamra2020localization}  &\rmark&\textbf{60.9}&267.4   &\textbf{89.0}&159.0&61.2&104.6&7.8&13.7  & 184.1 & 258.3\\
     \textbf{Ours*}&\rmark &66.6&\textbf{253.6}   &\textbf{89.0}&\textbf{153.5}& \textbf{57.0}&\textbf{103.4} &\textbf{6.9}&~\textbf{11.8}  & \textbf{171.4} & \textbf{233.1}\\
     \bottomrule
\end{tabular}}
\end{table*}

\subsection{Crowd counting} 
In this work, we mainly focus on the crowd localization task, while the counting result can also be easily obtained since the total count is equal to the number of local maxima. Tab.~\ref{tab:NWPU_counting}, and Tab.~\ref{tab:qnrf_A_B_counting} show the quantitative counting results of our method and state-of-the-art methods. 

\textbf{Compared with the localization-based methods}, which can provide the position information, our method significantly outperforms the state-of-the-art localization-based method SCALNet~\cite{wang2021dense} on the NWPU-Crowd (test set) by a signiﬁcant margin of 27.4 MSE. Our method also obtains the best performance on UCF-QNRF, ShanghaiTech Part A, Part B, and UCF\_CC\_50 datasets. For the JHU-Crowd++ dataset, the proposed method achieves SOTA performance in MSE and comparable performance in MAE. It indicates that the proposed method can cope with both sparse crowd scenes and dense crowd scenes.

\begin{figure*}[t]
\centering
\includegraphics[width=1\linewidth]{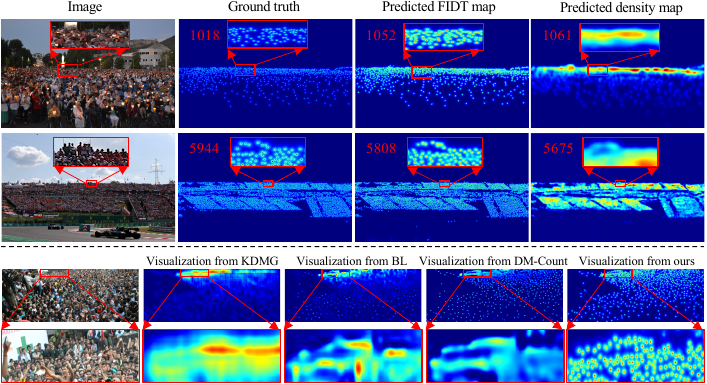}
\caption{Row 1 and row 2, from left to right, there are images, ground truth, predicted FIDT maps, and predicted density maps (trained with the same network). It can be seen that the heads of the predicted FIDT maps are distinguishable, and the predicted density maps show severe overlaps. Row 3 and row 4 compare the predicted visualizations from KDMG~\cite{wan2020kernel}, BL~\cite{ma2019bayesian}, DM-Count~\cite{wang2020distribution}, and ours. Our method provides precise location information in the dense region (red box).}
\centering
\label{fig:visiual}
\end{figure*}

\textbf{Compared with the regression-based methods}. Although it is not fair to compare localization-based counting methods and density-map regression-based counting methods, our method still outperforms all density map regression-based methods on NWPU-Crowd, JHU-Crowd++, and UCF\_CC\_50 datasets. Meanwhile, the proposed method achieves comparable performance on UCF-QNRF, ShanghaiTech Part A, and Part B datasets. To intuitively demonstrate the difference between FIDT maps and density maps, we provide the predicted FIDT maps and density maps visualization (the density maps are trained with the same network), as shown in Fig.~\ref{fig:visiual} (row 1 and row 2). We can see that the predicted density maps lose the position information and show severe overlaps in dense regions. However, the predicted FIDT maps provide an almost accurate location for each individual, even in the extremely dense scene. On row 3 and row 4 of Fig.~\ref{fig:visiual}, it compares the predicted visualizations from KDMG~\cite{wan2020kernel}, BL~\cite{ma2019bayesian}, DM-Count~\cite{wang2020distribution}, and ours. We can see that the DM-Count~\cite{wang2020distribution} and our method provide clear position information in the sparse region, while only our method can provide precise location information in the dense region (red box). 

\subsection{Evaluation on vehicle dataset}
A robustness algorithm should easily generalize to similar tasks (\textit{e.g.,} vehicle localization and counting). Thus, following previous localization methods~\cite{liu2019point,sam2020locate}, we evaluate the generalization capability of the proposed method on the TRANCOS~\cite{guerrero2015extremely} dataset for vehicle counting. We adopt the Grid Average Mean Absolute Error (GAME)~\cite{guerrero2015extremely} as the evaluation metric for vehicle counting, deﬁned as: 
\begin{equation}
GAME(L)=\frac{1}{N}\sum_{i=1}^{N}\left ( \sum_{l=1}^{4^{L}}\left | P_{i}^{l} -G_{i}^{l}\right | \right ),
\label{eq:GAME (L)}
\end{equation}
which splits an input image into $4^{L}$ non-overlapping sub-regions. $N$ is the number of the testing images, $P_{i}$ and $G_{i}$ are the predicted and ground truth count of the $i$-th image, respectively.
Tab.~\ref{tab:vehicle} compares the GAME metric of the proposed method and the state-of-the-art localization-based methods. Specifically, the proposed method achieves the best performance on GAME(0), GAME(1), and GAME(2) and obtains comparable performance on GAME(3). It means that the proposed method not only achieves accurate global predictions but also has well localization performance.

\subsection{Ablation Study} 

\begin{figure}[t]
	\begin{center}
		\includegraphics[width=0.96\linewidth]{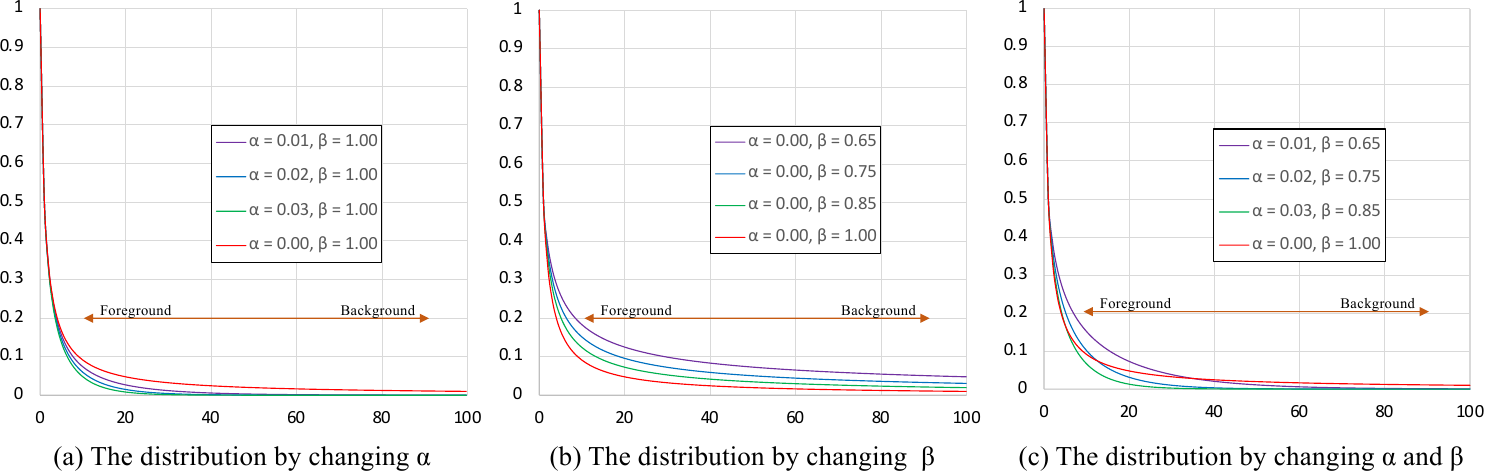}
	\end{center}
	\caption{The effect of changing $\alpha$ and $\beta$ on the distribution of FIDT map. }
	\label{fig:changing_alpha_beta}
\end{figure}

\textbf{Analysis of the FIDT map.} 
To understand the FIDT map better, we analyze the distribution of the FIDT map by using different $\alpha$ and $\beta$. Only changing the $\alpha$ (resp. $\beta$), as shown in Fig.~\ref{fig:changing_alpha_beta}, as $\alpha$ (resp. $\beta$) increasing (resp. decreasing), the response of FIDT map shows faster (resp. slower) decay in both foreground and background. As discussed in Sec.~\ref{sec:fidt}, the decay should be slower away from the head, and the response of the background should quickly close to 0. Thus, we set $\alpha$ = 0.02 and $\beta$ = 0.75 in all experiments, and we also report the various $\alpha$ and $\beta$ settings experiments in Tab.~\ref{tab:effectiveness_alpha}. The following ablation study will provide experiments of choosing 0.02 and 0.75 in Eq.~\ref{eq:fidt_map}.

\textbf{Effectiveness of I-SSIM loss.} 
In this section, we explore the advantage of the proposed I-SSIM loss. 
Based on Tab.~\ref{tab:I-SSIM}, we make the following observations: (1) Adding the traditional global SSIM loss can bring improvement. (2) The proposed I-SSIM loss achieves further improvement in terms of localization and counting,  mainly because the I-SSIM loss can further optimize the structure information of the predicted FIDT map to find local maxima better and repress the false local maxima in the background.

\begin{table}[t]
\small
\setlength{\tabcolsep}{1mm}
\centering
\caption{Quantitative comparison of vehicles counting on the TRANCOS~\cite{guerrero2015extremely} dataset. $\dag$ represents that the networks are trained by ourselves.}
\label{tab:vehicle}
\begin{tabular}{ lccccc }
\toprule
Method &GAME(0)&GAME(1)&GAME(2)&GAME(3)\\
\cmidrule(r){1-1} \cmidrule(r){2-5}
PSDDN~\cite{liu2019point}&4.79& 5.43&6.68&8.40\\
LSC-CNN~\cite{sam2020locate} &4.60&5.40&6.90&\textbf{8.30}\\
AutoScale~\cite{xu2022autoscale} &2.88&4.97&6.64&9.73\\
Crowd-SDNet\dag~\cite{wang2021self} &3.82&5.27&7.72&10.11\\
TopoCount\dag~\cite{abousamra2020localization} &3.42&4.76&6.51&8.55\\
\textbf{Ours}&\textbf{2.25} & \textbf{3.91} &\textbf{5.66}&8.36\\
\bottomrule
\end{tabular}
\end{table}

\begin{table}[t]
\small
\setlength{\tabcolsep}{1.5mm}
\centering
\caption{The results of different $\alpha$ and $\beta$ setting on Part A. It is noteworthy that IDT map (Eq.~\ref{eq:idt_map}) means the $\alpha$ = 0 and $\beta$ = 1.}
\label{tab:effectiveness_alpha}

\begin{tabular}{ cccccccc }
\toprule
\multirow{2}{*}{Method}& \multirow{2}{*}{$\alpha $}&\multirow{2}{*}{$\beta$}&\multicolumn{3}{c}{Localization ($\sigma$ = 8)}&\multicolumn{2}{c}{Counting}\\
 \cmidrule(r){4-6} \cmidrule(r){7-8}
 &&& P(\%) & R(\%)& F(\%) &MAE & MSE\\
\cmidrule(r){1-1} \cmidrule(r){2-3} \cmidrule(r){4-6} \cmidrule(r){7-8}
 IDT &0.00 & 1.00 &75.6\% &74.6\% &75.1\% & 61.8 & 109.6  \\
 FIDT &0.01 & 0.65  &76.6\% &76.0\% & 76.3\%& 60.5 & 107.4 \\
 FIDT &0.02 & 0.75 &\textbf{78.2\%}&\textbf{77.0\%}&\textcolor{black}{\textbf{77.6}\%}& \textbf{57.0} & \textbf{103.4}   \\
 FIDT &0.03 & 0.85 &77.0\% & 76.8\% & 76.9\%& 58.3 & 106.6   \\
\bottomrule
\end{tabular}
\end{table}

\begin{figure}[t]
	\begin{center}
	\includegraphics[width=0.96\linewidth]{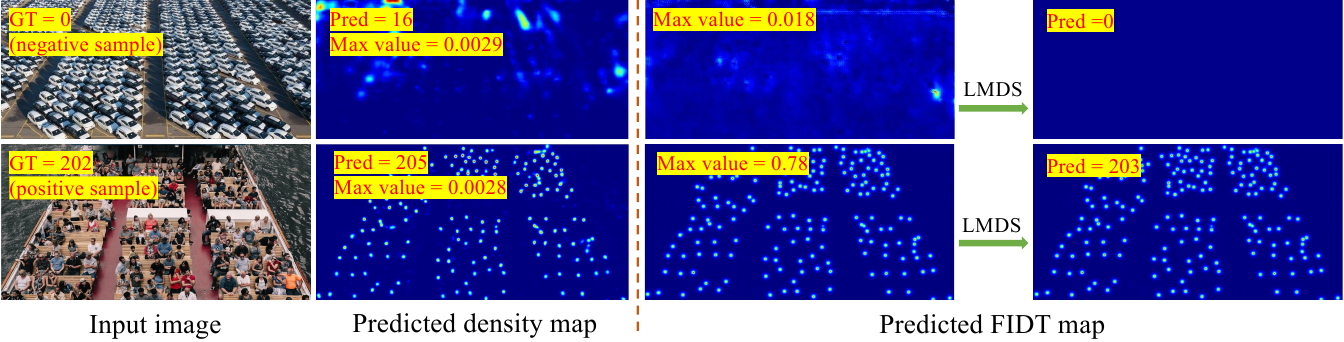}
	\end{center}
	\caption{
Row 1 and row 2 are negative and positive samples, respectively. In the density maps, the max pixel value between positive and negative samples is very similar. In the FIDT maps, the negative and positive samples present a significant difference in max pixel value. 
	}
	\label{fig:negative_fidt}
\end{figure}

\begin{table}
\small
\setlength{\tabcolsep}{1.5mm}
\centering
\caption{The effectiveness of the proposed I-SSIM loss on Part A.}
\label{tab:I-SSIM}
\begin{tabular}{ cccccc }
 \toprule
 {\multirow{2}{*}{Method}} & \multicolumn{3}{c}{Localization ($\sigma$ = 8)}  & \multicolumn{2}{c}{Counting}\\
\cmidrule(r){2-4} \cmidrule(r){5-6}
  & P(\%) & R(\%)& F(\%)&MAE & MSE   \\
\cmidrule(r){1-1} \cmidrule(r){2-4} \cmidrule(r){5-6}
  L2&73.5\% & 76.3\%  &74.9\%&62.1&108.8 \\
  L2 + SSIM & 76.8\% &76.6\% &76.7\%&59.3&106.5\\
  L2 + I-SSIM \textbf{(ours)} & \textbf{78.2\%}&\textbf{77.0\%}&\textcolor{black}{\textbf{77.6}\%}&\textbf{57.0}&\textbf{103.4} \\
\bottomrule
\end{tabular}
\end{table}

\begin{table}
\small
\centering
\caption{The influence of the independent instance region size of I-SSIM loss on Part A dataset.}
\label{tab:influence_region_size}
\begin{tabular}{ cccccc }
\toprule
 \multirow{2}{*}{Region Size}&\multicolumn{3}{c}{Localization ($\sigma$ = 8)}&\multicolumn{2}{c}{Counting}\\
\cmidrule(r){2-4} \cmidrule(r){5-6}
 & P(\%) & R(\%)& F(\%) &MAE & MSE\\
\cmidrule(r){1-1} \cmidrule(r){2-4} \cmidrule(r){5-6}
 $20\times20$  & 78.0\%  & 76.4\% & 77.2\% & 58.1 & 104.4  \\
 $30\times30$  & \textbf{78.2\%} & \textbf{77.0\%} & \textbf{77.6\%} & \textbf{57.0} & \textbf{103.4}  \\
 $40\times40$  & 77.4\% & 76.4\% & 76.9\% & 58.6 & 105.9   \\
 \bottomrule
\end{tabular}
\end{table}

\begin{table}
\setlength{\tabcolsep}{10pt}
\small
\centering
\caption{The ablation study on the threshold $T_a$.}
\label{tab:adaptive_threshold}
\begin{tabular}{ cccc }
\toprule
Threshold value & Adaptive & MAE & MSE \\
\midrule
$50/255$ &\xmark & 95.6&169.9\\
$70/255$  &\xmark & 92.1&167.0\\
$90/255$ &\xmark & 80.6&152.3\\
$100/255$ &\xmark & 115.3&206.4\\
$110/255$  &\xmark & 122.6&233.5\\
\midrule
$90/255 \times max(M)$ &\rmark & 60.4&105.3\\
$100/255 \times max(M)$  &\rmark &\textbf{57.0}&\textbf{103.4}\\
$110/255 \times max(M)$  &\rmark & 58.1&107.1\\
\bottomrule
\end{tabular}
\end{table}

We further ablate the influence of the independent instance region sizes of I-SSIM loss, as shown in Tab.~\ref{tab:influence_region_size}. Larger region sizes may contain too much background (without structure information), leading to excess false local maxima in the background. Smaller region sizes may not involve the entire independent head, which can not effectively enhance the structure information of the local maxima (head region). 
Based on the experiments, we use $30\times30$ as the region size for all datasets, and it works well.  

\textbf{Analysis of $T_a$.}
On the proposed post-processing, LMDS, the $T_a$ is used to choose the positive points. Its value is adaptive, which is set to $\frac{100}{255} \times \max(M)$, where $max(M)$ is the max value of the predicted FIDT map. As shown in Tab.~\ref{tab:adaptive_threshold}, using fixed thresholds is worse than the adaptive threshold since the local-maxima pixel values of different predicted FIDT maps are not the same. This inspires us to utilize adaptive threshold. Using large adaptive thresholds will filter out the true-positive points, and small will reserve some false-positive points. Hence, we choose the $\frac{100}{255} \times \max(M)$ as an adaptive threshold for all datasets. 

\begin{table}
\small
\centering
\caption{The results of S0 and S4 on NWPU-Crowd test set.}
\label{tab:negative_samples}
\setlength{\tabcolsep}{3mm}
\begin{tabular}{ ccccc }
\toprule
  {\multirow{2}{*}{Method}}&\multicolumn{2}{c}{S$0$-level}&\multicolumn{2}{c}{S$4$-level}\\
  \cmidrule(r){2-3} \cmidrule(r){4-5}
  &MAE & MSE & MAE & MSE\\
\cmidrule(r){1-1} \cmidrule(r){2-3} \cmidrule(r){4-5}
KDMG~\cite{wan2020kernel} &77.3&303.0&2777.9&3521.8\\
DM-Count~\cite{wang2020distribution} &146.7 & 736.1 &2075.8&2895.2 \\
NoisyCC~\cite{wan2020modeling} &218.7 & 1415.6 &2572.5& 3414.9\\
SCALNet~\cite{wang2021dense} &92.0&479.3&1775.3&2676.4\\
 \textbf{Ours} &\textbf{21.6} & \textbf{129.3}&\textbf{1645.4}&\textbf{2288.2}   \\
\bottomrule
\end{tabular}
\end{table}

\textbf{Robustness on negative and dense scenes.} 
The S0 of NWPU-Crowd consists of some ``dense fake humans'' (\textit{e.g.,} Terra-Cotta Warriors), called negative samples. In contrast, S4 means the extremely dense crowd scenes, containing more than 5,000 persons. Thus, the S0 and S4 are usually adopted to evaluate the model's robustness~\cite{gao2020nwpu}. Tab.~\ref{tab:negative_samples} lists the results of some popular methods on the NWPU-Crowd’s negative samples and extremely dense scenes. As expected, the proposed method achieves the lowest counting error, reporting superior robustness. As shown in Fig.~\ref{fig:negative_fidt}, the maximum pixel value between negative and positive samples is similar for density maps while it is distinct for the FIDT maps. Given a predicted FIDT map, if its maximum pixel value is smaller than~threshold $T_f$, the LMDS will regard the input as a negative sample and set the counting result as 0, as illustrated in Algorithm~\ref{alg:extract_position}.

\begin{figure}[t]
\centering
\includegraphics[width=0.96\linewidth]{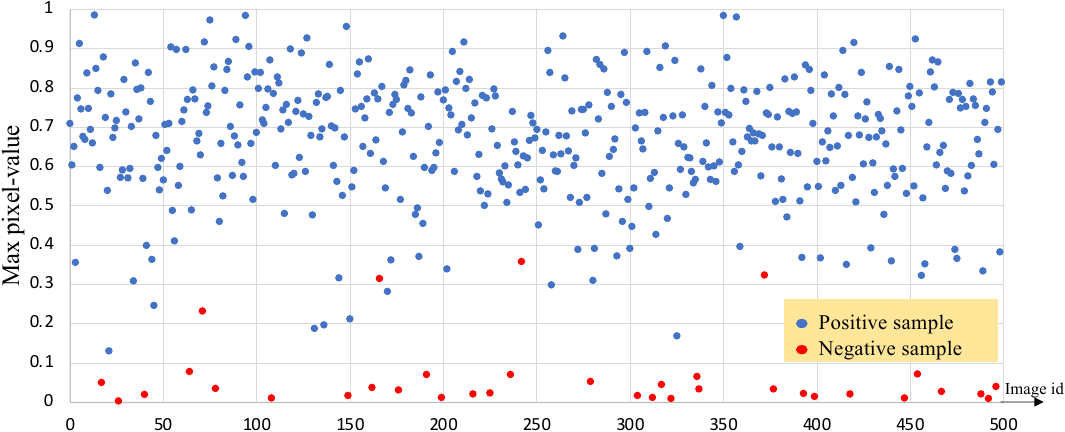}
\caption{The max pixel-value distribution of predicted FIDT maps (NWPU-Crowd validation set).}
\centering
\label{fig:max_pixel}
\end{figure}

We set the $T_f$ as 0.1 according to the statistics. Specifically, we give the max pixel-value of each predicted FIDT map based on NWPU-Crowd (validation set, including 500 images), as shown in Fig.~\ref{fig:max_pixel}. We can observe that all positive samples' value is much bigger than 0.1, and most negative samples are smaller than 0.1. Thus, the threshold $T_f$ set as 0.1 is reasonable.

\textbf{Generalization on different regressors.} 
In this section, to demonstrate the proposed FIDT map can be generalized to different regressors, we implement the CSRNET~\cite{li2018csrnet}, BL~\cite{ma2019bayesian} with the FIDT maps on the NWPU-Crowd dataset (validation set). Besides, we add a FPN into the CSRNET~\cite{li2018csrnet} and BL~\cite{ma2019bayesian} to capture rich spatial context. The quantitative results are listed in Tab.~\ref{tab:different_regressor}, where we can observe that using the FIDT map can realize the localization task, and the counting performance is competitive compared with the density map. Notability, the experiments of FIDT maps only adopt the L2 loss, and the image scaling strategy is the same as ~\cite{gao2020nwpu}. The results indicate that the FIDT map is suitable for the crowd localization task.

\begin{table}[t]
    \footnotesize
    \setlength{\tabcolsep}{2mm}
	\centering
	\caption{The results of different regressors on the NWPU-Crowd~\cite{gao2020nwpu} (validation set) dataset.}
	\label{tab:different_regressor}
    \resizebox{0.96\linewidth}{!}{
	\begin{tabular}{lccccccc}
		\toprule
		\multirow{4}{*}{Method} & \multicolumn{2}{c}{\multirow{2}*{\tabincell{c}{Density map\\(counting by integration)}}}
		& \multicolumn{5}{c}{\multirow{2}*{\tabincell{c}{FIDT map\\(counting by localization)}}} \\
		&\multicolumn{2}{c}{ }&\multicolumn{5}{c}{}\\
 		\cmidrule(r){2-3} \cmidrule(r){4-8}
 		&MAE&MSE&MAE&MSE &P(\%) & R(\%)&F(\%)  \\
		\cmidrule(r){1-1} \cmidrule(r){2-3} \cmidrule(r){4-8}
        CSRNET~\cite{li2018csrnet} &104.9 & 433.5&100.6&464.3 &68.8\% & 66.2\% &67.5\%\\
        CSRNET~\cite{li2018csrnet} + FPN &95.5 & 450.8 &70.6&369.7 &73.9\% & 69.8\% & 71.8\%\\ 
        BL~\cite{ma2019bayesian} &93.6& 470.4 &89.7&446.6 & 69.9\% & 68.9\% & 69.4\%\\  
 		BL~\cite{ma2019bayesian} + FPN &85.4& 412.9 &65.7&259.2&77.8\%&70.0\%&73.7\% \\  
		\bottomrule
	\end{tabular}}
\end{table}%

\begin{table}
\small
    \setlength{\tabcolsep}{2mm}
\centering
\caption{The comparisons of complexity. The F-measure is from the NWPU-Crowd benchmark (test set).}
\label{tab:run_time}
\begin{tabular}{ ccccc }
\toprule
Method & MACs (G) & Inference speed & F-measure \\
\cmidrule(r){1-1} \cmidrule(r){2-4}
LSC-CNN~\cite{sam2020locate}   & 1244.3 &2.6 FPS& -\\
AutoScale~\cite{xu2022autoscale}  & 1074.6 &5.7 FPS& 62.0\% \\
Crowd-SDNet\footnote{We try our best to calculate the MACs of Crowd-SDNet, but the official code relies on the old version Keras, which is hard to obtain the MACs.}~\cite{wang2021self} & - &  0.8 FPS & 63.7\% \\
GL~\cite{wan2021generalized} & \textbf{324.6} & \textbf{20.3} FPS & 66.0\% \\
TopoCount~\cite{wang2020distribution}  & 797.2 & 9.4 FPS& 69.1\% \\
\textbf{Ours} & 426.7& 7.1 FPS & \textbf{75.5}\% \\
\bottomrule
\end{tabular}
\end{table}

\subsection{Limitation}

The main limitation is that the proposed method inference will be slower than some real-time methods~\cite{wan2021generalized}. As shown in Tab.~\ref{tab:run_time}, we report the Multiply-Accumulate Operations (MACs) and Frames Per Second (FPS) to analyze the complexity. All methods are evaluated on the official code with a size of 768 $\times$ 1024 image, and the GPU device is NVIDIA RTX 3090. Although our method achieves the second MACs and the third FPS, there is still a lot of room for improvement. In the future, we are interested in extending our method for real-time.

\section{Conclusion}
In this paper, we present a novel label named FIDT map, designed to cope with the crowd localization task. The proposed FIDT map is a non-overlap map, which utilizes local maxima to represent the head‘s center. To extract the corresponding individual center, a Local-Maxima-Detection-Strategy (LMDS) is proposed. Besides, we introduce a novel I-SSIM loss to make the model tend to focus on the foreground regions, improving the structure information of local maxima. By performing experiments on six publicly available datasets, we demonstrate that the proposed method achieves state-of-the-art localization performance and shows superior robustness for the negative samples and extreme scenes. We hope the community switches from the density map regression to FIDT map regression for more practical.

{\small
\bibliographystyle{ieee_fullname}
\bibliography{egbib}
}

\begin{IEEEbiography}[{\includegraphics[width=1in,height=1.25in,clip]{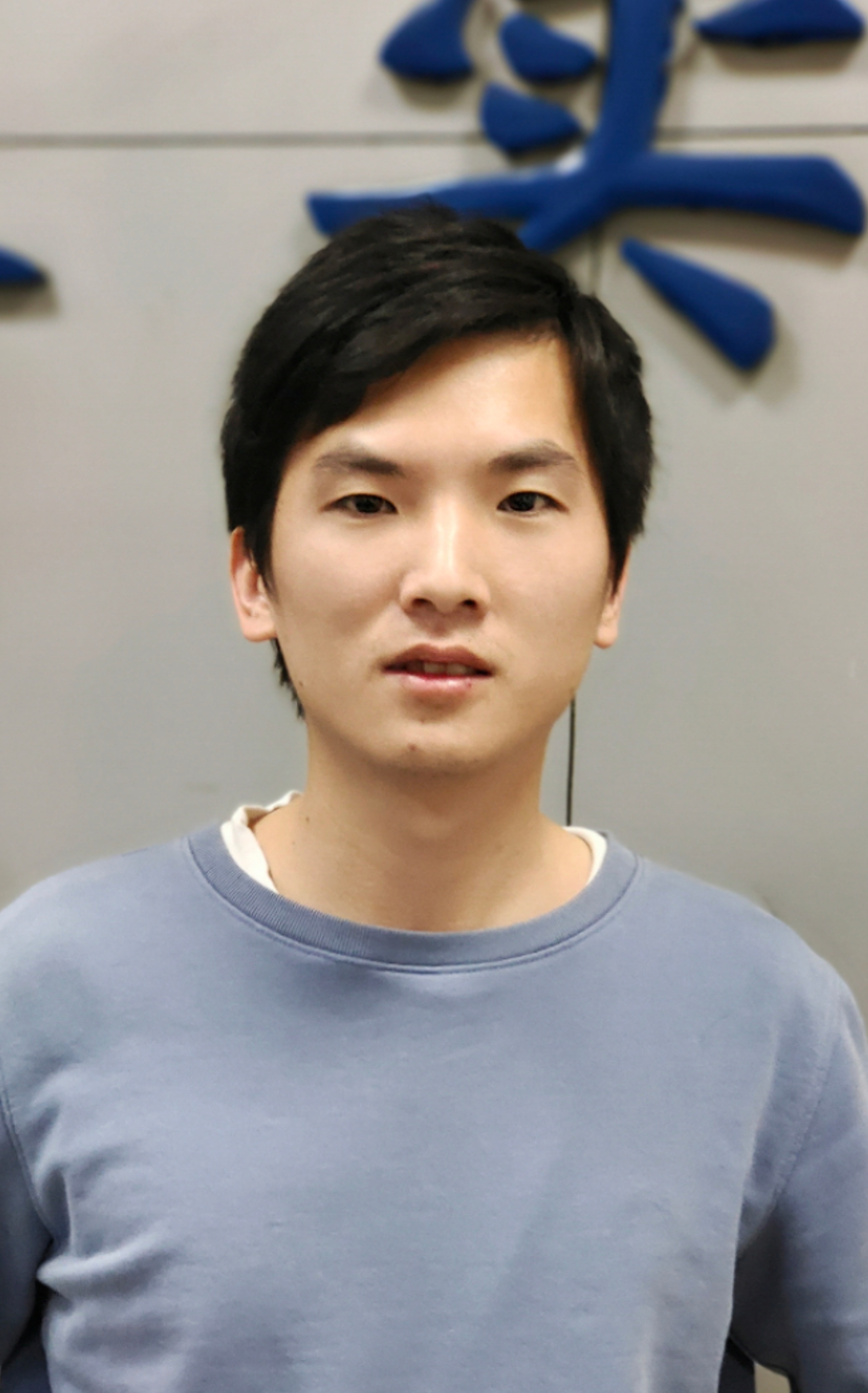}}]{Dingkang Liang}  is currently working towards the Ph.D. degree in School of Artificial Intelligence and Automation from the Huazhong University of Science and Technology, Wuhan, China. He has served as a reviewer for several top journals and conferences such as TPAMI, TIP, CVPR, ICCV, and ECCV. His research interests include computer vision, especially for crowd analysis and 3D object detection.
\end{IEEEbiography}

\begin{IEEEbiography}[{\includegraphics[width=1in,height=1.25in,clip,keepaspectratio]{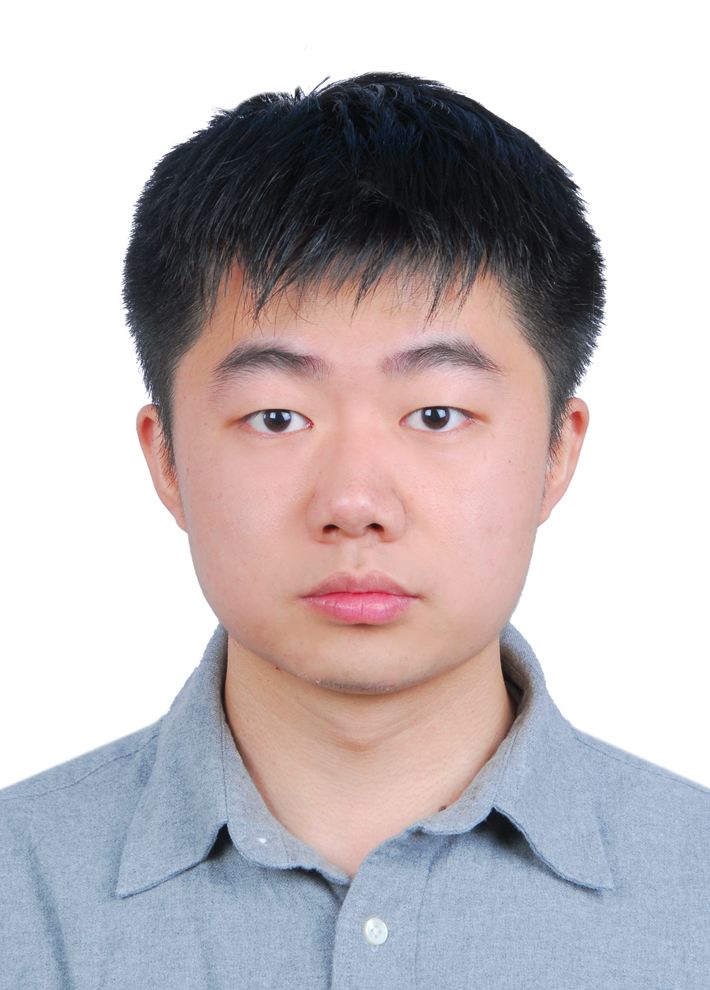}}]{Wei Xu} is currently working towards the M.S. degree in information and communication engineering from Beijing University of Posts and Telecommunications, Beijing, China. His research interests include crowd analysis, face reconstruction, and object detection.
\end{IEEEbiography}

\begin{IEEEbiography}[{\includegraphics[width=1in,height=1.25in,clip,keepaspectratio]{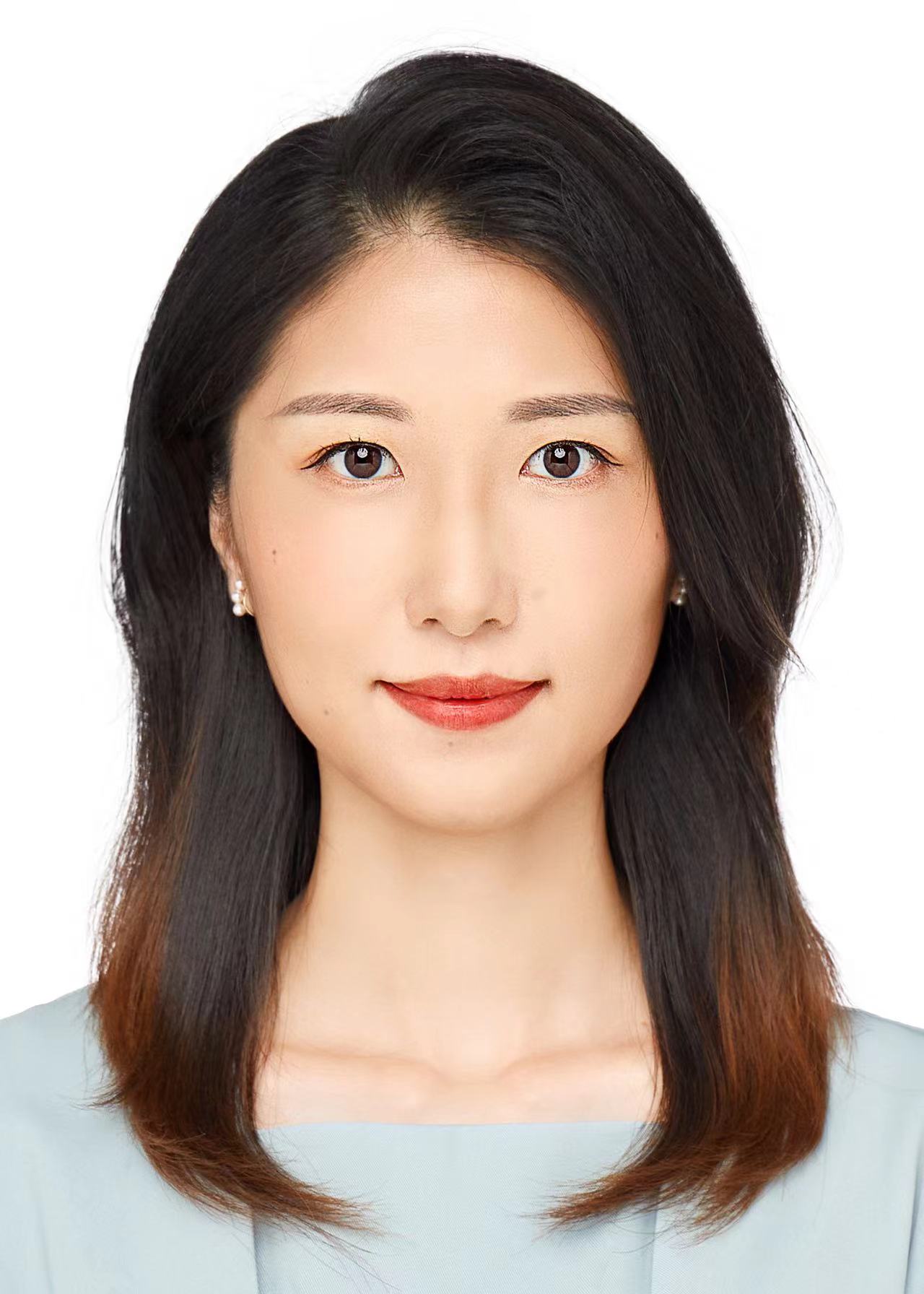}}]{Yingying Zhu} received the bachelor's degree and Ph.D. degree in Electronics and Information Engineering from Huazhong University of Science and Technology (HUST), Wuhan, P.R. China in 2018. She joined the  Huazhong University of Science and Technology (HUST), Wuhan, as an Engineer from 2021 to now. Her research interests include computer vision and machine learning. 
\end{IEEEbiography}

\begin{IEEEbiography}[{\includegraphics[width=1in,height=1.25in,clip,keepaspectratio]{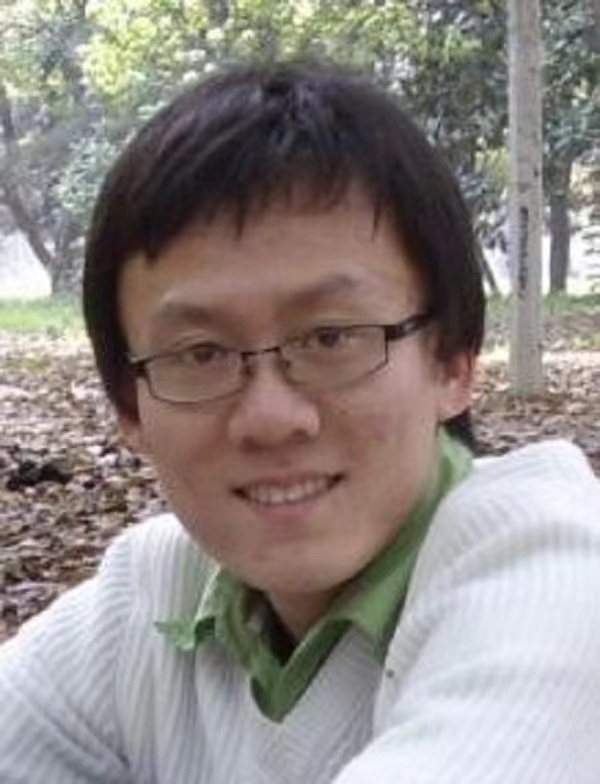}}]{Yu Zhou} received the M.S. and Ph.D. degrees both in Electronics and Information Engineering from Huazhong University of Science and Technology (HUST), Wuhan, P.R. China in 2010, and 2014, respectively. In 2014, he joined the Beijing University of Posts and Telecommunications (BUPT), Beijing, as a Postdoctoral Researcher from 2014 to 2016, an Assistant Professor from 2016 to 2018. He is currently an Associate Professor with the School of Electronic Information and Communications, HUST. His research interests include computer vision and automatic drive. 
\end{IEEEbiography}

\end{document}